%% file: main.tex
\documentclass[a4paper, 10pt, conference]{ieeeconf}      

\IEEEoverridecommandlockouts                              

\usepackage{times}
\usepackage{epsfig}
\usepackage{graphicx}
\usepackage{amsmath}
\usepackage{bm}
\usepackage{amssymb}

\usepackage{hyperref}
\usepackage{tabularx}
\usepackage{amsmath}
\usepackage{siunitx}
\usepackage{booktabs}
\usepackage{multirow}
\usepackage[table,xcdraw]{xcolor}
\usepackage{float}
\usepackage[normalem]{ulem}
\useunder{\uline}{\ul}{}
\newcommand{\etal}{et al.}
\definecolor{lightergray}{rgb}{0.8, 0.8, 0.8}
\usepackage{cite}

\title{\LARGE \bf
Monocular Vision based Crowdsourced 3D Traffic Sign Positioning with Unknown Camera Intrinsics and Distortion Coefficients
}

\author{Hemang Chawla, Matti Jukola, Elahe Arani, and Bahram Zonooz
\thanks{All authors are with Advanced Research Lab, NavInfo Europe, Eindhoven, The Netherlands. \tt\small \{hemang.chawla, matti.jukola, elahe.arani, b.yoosefizonooz\}@navinfo.eu }
}

\begin{document}

\input{ieee_copyright}
\newpage


\maketitle
\thispagestyle{empty}
\pagestyle{empty}

\begin{abstract}
\input{abstract.tex}
\end{abstract}

\input{introduction.tex}

\input{related_work.tex}

\input{method.tex}
\input{experiments.tex}
\input{conclusion.tex}

\addtolength{\textheight}{-2cm}   

\bibliographystyle{IEEEtran}
\bibliography{IEEEabrv,refbib}

\end{document}

%% file: ieee_copyright.tex
\thispagestyle{empty}
\onecolumn
\noindent
\large
\textbf{This paper has been accepted for publication in the proceedings of 
\textit{2020 IEEE 23rd International Conference on Intelligent Transportation Systems (ITSC)}, Rhodes, Greece, September 20-23, 2020.}

\bigskip\bigskip
\noindent
\large
IEEE Copyright notice:\\\\
\normalsize
\copyright 2020 IEEE. Personal use of this material is permitted. Permission from IEEE must be obtained for all other uses, in any current or future media, including reprinting /republishing this material for advertising or promotional purposes, creating new collective  works,  for  resale  or  redistribution  to  servers  or  lists,  or  reuse  of  any  copyrighted  component  of  this  work  in  other works.

\bigskip\bigskip
\noindent
\large
Cite as:\\\\
\noindent\fbox{%
    \parbox{\textwidth}{%
    \noindent
    \normalsize
       H. Chawla, M. Jukola, E. Arani, and B. Zonooz, ``Monocular Vision based Crowdsourced 3D Traffic Sign Positioning with Unknown Camera Intrinsics and Distortion Coefficients," \textit{2020 IEEE 23rd International Conference on Intelligent Transportation Systems (ITSC)}, Rhodes, Greece, IEEE (in press), 2020.
    }%
}

\bigskip\bigskip
\noindent
\large
\textsc{Bib}\TeX:\\\\
\noindent\fbox{%
    \parbox{\textwidth}{%
    \noindent
    \normalsize
    \texttt{\noindent @inproceedings\{chawla2020monocular,\\
        author=\{H. \{Chawla\}, and M. \{Jukola\}, and E. \{Arani\}, and B. \{Zonooz\}\},\\
        booktitle=\{2020 IEEE 23rd International Conference on Intelligent Transportation Systems (ITSC)\}, \\
        title=\{Monocular Vision based Crowdsourced 3D Traffic Sign Positioning with Unknown  Camera  Intrinsics  and  Distortion  Coefficients\}, \\
        location=\{Rhodes, Greece\},\\
        publisher=\{IEEE (in press)\},\\
        year=\{2020\}     
        }%
    }%
}

\normalsize
\twocolumn

%% file: abstract.tex
Autonomous vehicles and driver assistance systems utilize maps of 3D semantic landmarks for improved decision making. However, scaling the mapping process as well as regularly updating such maps come with a huge cost. Crowdsourced mapping of these landmarks such as traffic sign positions provides an appealing alternative. The state-of-the-art approaches to crowdsourced mapping use ground truth camera parameters, which may not always be known or may change over time. In this work, we demonstrate an approach to computing 3D traffic sign positions without knowing the camera focal lengths, principal point, and distortion coefficients a priori. We validate our proposed approach on a public dataset of traffic signs in KITTI. Using only a monocular color camera and GPS, we achieve an average single journey relative and absolute positioning accuracy of  \SI[detect-weight=true, detect-family=true, mode=text]{0.26}{\m} and \SI[detect-weight=true, detect-family=true,mode=text]{1.38}{\m}, respectively.

%% file: introduction.tex
\section{Introduction}
Recent developments in computer vision, mapping, and localization technology have led to major progress of modern autonomous driving prototypes and driver assistance systems. 
For accurate and safe action planning and decision making, landmark-based maps describing 3D geometry of road features, traffic signage, lane intersections, and other semantic objects are necessary. However, creating such maps is costly due to the use of dedicated collection vehicles with Light Detection and Ranging (LiDAR) sensors, stereo cameras, Inertial Measurement Units (IMU), Global Positioning System (GPS), wheel odometers, and radars, fitted onto them~\cite{jiao2018machine}, thereby limiting their scope. It is also desired that short term changes (for instance due to road maintenance), and long term changes in road structure are reflected in these maps. Therefore, using dedicated mapping equipment is a bottleneck for regular creation and update of these 3D maps.

\begin{figure}[t]
\begin{center}
  \includegraphics[width=0.95\linewidth]{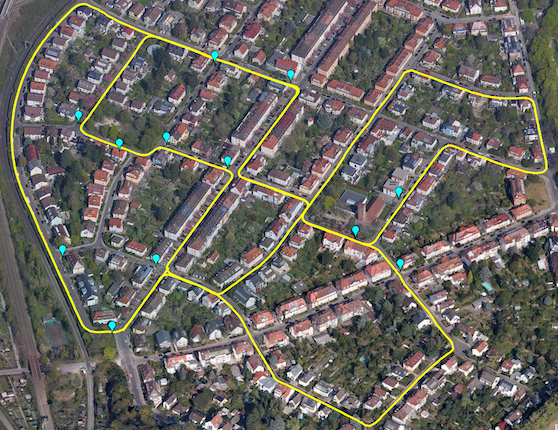}
\end{center}
  \caption{3D traffic sign triangulation in Germany without prior knowledge of camera parameters. The estimated signs are shown in cyan. The computed path of the vehicle used for triangulation is depicted in yellow.}
\label{fig:self_calib_sensitivity_oat}
\end{figure}

Crowdsourced maps built using a limited number of consumer-grade sensors provide an appealing solution to this problem. Monocular color cameras and GPS are easily available sensors for constructing crowdsourced maps. However, the calibration parameters of cameras used in such a system may be unknown or change over time. The state-of-the-art solution to crowdsourced mapping utilizes GPS, IMU, and monocular color cameras assuming known camera intrinsics and distortion coefficients \cite{dabeer2017end}. 

Therefore, to expand the scope of crowdsourced mapping, it is required to perform camera self-calibration followed by monocular ego-motion estimation and triangulation of the landmarks. 
Over the years, multiple approaches have been proposed to estimate the camera parameters without the use of external calibration objects like a checkerboard. Using two or more views of the scene, the distortion parameters \cite{byrod2008fast,kukelova2007minimal,kukelova2015efficient} and the 
focal lengths \cite{bocquillon2007constant,gherardi2010practical,hartley1993extraction,sturm2001focal} are estimated. 
However, calibration of the principal point is an ill-posed problem \cite{de1998self}, hence it is often fixed at the image center. 
Structure from motion (SfM) reconstruction has also been applied to estimate and optimize the camera parameters \cite{pollefeys2008detailed, schonberger2016structure, schoenberger2016mvs}. Even though self-calibration is pertinent to crowdsource 3D traffic sign positions from distorted image sequences with unknown camera calibration, its utility has not been analyzed until now. 

In this work, we demonstrate crowdsourced mapping focused on the positioning of 3D traffic signs, given their importance in the safety of autonomous driving systems, as well as the maintenance of traffic device inventory.  We propose a framework to estimate the traffic signs positions from a sequence of distorted images captured from a camera with unknown parameters, and corresponding GPS positions. 
Furthermore, we analyze the sensitivity of 3D traffic sign position triangulation to the accuracy of the camera focal lengths, principal point, and distortion coefficients.

\begin{figure*}[thbp!]
\begin{center}
    \includegraphics[width=0.9\linewidth]{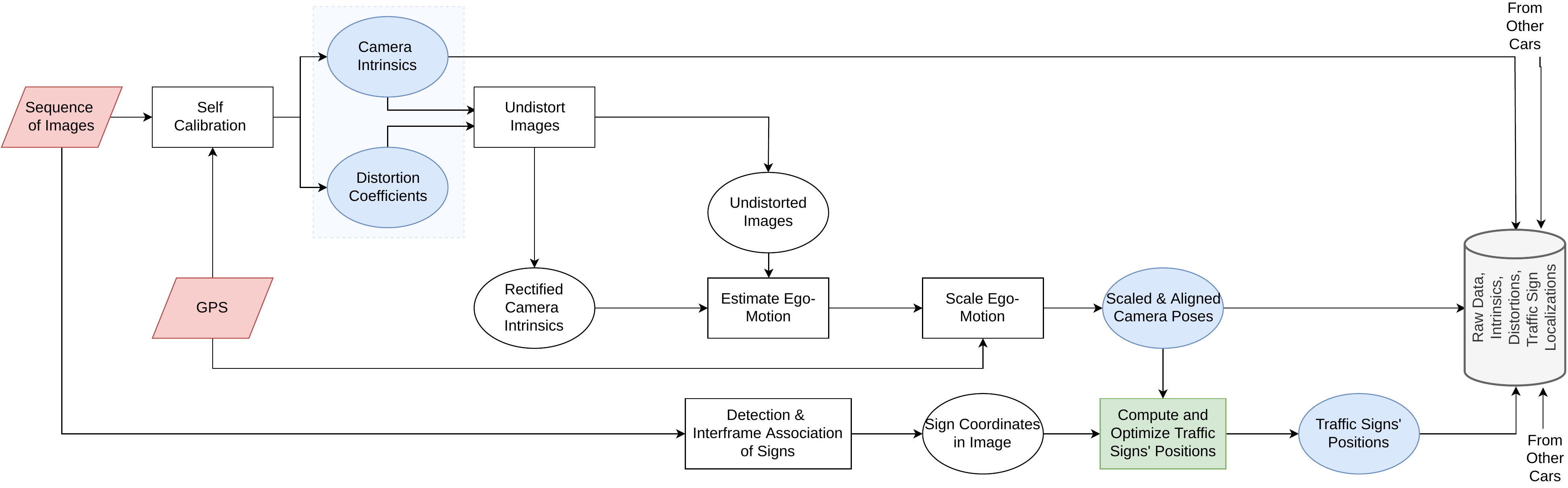}
\end{center}
   \caption{Single journey 3D traffic sign positioning framework without prior knowledge of camera intrinsics and distortion coefficients. The pink components represent inputs to the framework. The blue components represent the outputs of the primary steps of the approach. The crowdsourced mapping system in grey depicts the traffic sign positioning data collected through different cars.}
\label{fig:SingleJourney}
\end{figure*}

%% file: related_work.tex
\section{Related Work}
One of the first attempts to localizing traffic signs was aimed at inventorying the road attributes on highways \cite{arnoul1996traffic}. Using the Kalman filter for tracking the detected traffic signs and estimating their 3D positions, the method was limited to static scenes with the collection vehicle moving at a maximum speed of 5 km/h. In contrast, Madeira \etal~\cite{madeira2005automatic} developed a mobile mapping system that estimated the positions of traffic signs through photogrammetric triangulation within a least-squares approach, given the vehicle position from GPS, IMU, and wheel odometry fusion. In order to include signs found in crowded locations, Benesova \etal~\cite{benesova2007mobile} proposed an alternative approach of triangulating traffic sign positions using dedicated hand held devices. For extending to a real-time use-case, an approximate method was also proposed \cite{krsak2011traffic}. Another real-time traffic sign positioning method was proposed by Welzel \etal~\cite{welzel2014accurate}, using only a monocular color camera and GPS. However, the ground truth size and height of traffic signs in each class were used for computing their 3D positions to an accuracy of \SI[mode=text]{1}{\meter}. Similarly, a method for mapping positions of traffic lights 
was proposed \cite{fairfield2011traffic}.
Recently, Dabeer \etal~\cite{dabeer2017end} presented a method for crowdsourcing 3D positions and orientations of traffic signs using low-cost sensors. They demonstrated a single journey average relative and absolute positioning accuracy of \SI[detect-weight=true, detect-family=true, mode=text]{46}{\cm} and \SI[mode=text,detect-weight=true, detect-family=true]{57}{\cm}, respectively. 
However, all of the aforementioned approaches either used dedicated collection hardware for computing traffic sign positions, or assumed known accurate camera focal lengths, principal point, and distortion parameters. 


%% file: method.tex
\section{Method}
\label{section:system_overview}
In this section, we describe our proposed framework for GPS and monocular camera based 3D traffic sign positioning, without assuming any prior knowledge of the camera parameters. Given a sequence of $n$ color images and corresponding GPS positions as input, we output a set of $m$ detected traffic signs with their corresponding classes, absolute positions as well as the relative positions for the frames in which the sign was detected.
An overview of the proposed approach is shown in Fig. \ref{fig:SingleJourney}.
Hereafter, we describe the steps of computing 3D positions of traffic signs detected in crowdsourced image sequences.  

\subsection{Camera Self-Calibration} 
\label{section:self_calibration}
Crowdsourced mapping without prior knowledge of camera intrinsics and distortion parameters necessitates camera self-calibration. We use the pinhole camera model with zero skew
\begin{equation}
    K = \begin{bmatrix}
    f_x & 0 & c_x \\
    0 & f_y & c_y \\
    0 & 0 & 1
    \end{bmatrix}
\end{equation}
and the polynomial radial distortion model with two parameters
\begin{equation}
    \begin{bmatrix}
    x_d \\
    y_d
    \end{bmatrix}
     =
     (1 + \lambda_1 r^2 + \lambda_2 r^4)
     \begin{bmatrix}
    x_u \\
    y_u
    \end{bmatrix}.
\end{equation}
$f_x$ and $f_y$ are the focal lengths in $x$ and $y$, the principal point is represented by $(c_x, c_y)$, and $\lambda_1, \lambda_2$ are the distortion coefficients. The distance from the principal point is given by $r$. 
In this work, we use Structure from Motion based Colmap \cite{schonberger2016structure} with Oriented FAST and Rotated BRIEF (ORB) features \cite{rublee2011orb} for self-calibration. Since self-calibration suffers from ambiguity for the case of pure translation \cite{steger2012estimating, wu2014critical} due to scene depth and distortion conflation, we use the sub-sequences where the vehicle is making a turn. The sub-sequences containing the turns are extracted through the Ramer-Douglas-Peucker (RDP) algorithm \cite{ramer1972iterative, douglas1973algorithms} by decimating the GPS trajectory into a similar curve with fewer points, where each point represents a turn. 
Thereafter, the calibration is performed in two steps. In the first step, it is assumed that $f_x = f_y$, the principal point $(c_x, c_y) = \tfrac{1}{2}\cdot(w,h)$, where $w$ and $h$ are the width and height of the images respectively. The distortion is modeled using only $\lambda_1$, while $\lambda_2 = 0$.
In the second step, the aforementioned restrictions are relaxed and all the parameters are optimized simultaneously. 


\subsection{Estimating Camera Ego-Motion} 
\label{section:ego_motion_and_depth}
After computing the camera intrinsics and the distortion parameters, the camera ego-motion needs to be estimated as shown in Fig. \ref{fig:SingleJourney}. For this step, the images are first undistorted using the estimated parameters, and the rectified camera matrix is calculated. Thereafter, we use state-of-the-art geometry based monocular approach ORB-SLAM \cite{mur2015orb} for camera ego-motion estimation. 
Since monocular ego-motion estimation is valid up to scale, we then use the GPS positions to scale the estimated trajectory using the Umeyama's algorithm \cite{umeyama1991least}.
Firstly the GPS positions are converted to metric coordinates under the Mercator assumption such that
\begin{equation}
    x = \cos \left(\dfrac{\pi \cdot \text{lat}_0}{180}\right) r_{earth} \dfrac{\pi \cdot \text{lon}
    }{180},
\end{equation}
\begin{equation}
    y = \cos \left(\dfrac{\pi \cdot \text{lat}_0}{180}\right) r_{earth} \log \left(\tan\dfrac{\pi \cdot (90 + \text{lat})}{360}\right),
\end{equation}
where $r_{earth} = \SI{6378137}{\m}$.
Thereafter, to scale and align the estimated camera positions ($t_{0,j} \forall j=1\dots n$) with the GPS positions ($g_{0,j} \forall j=1\dots n$), a similarity transformation, (rotation $R$, translation $t$, and scale $s$) is computed minimizing the mean squared error 
\begin{equation}
\label{eq:umeyama}
    \varepsilon(R, t, s) = \dfrac{1}{n} \sum_{j=1}^{n} \lVert g_{0,j} - (sRt_{0,j} + t) \rVert^2 
\end{equation}
between them.
The scaled and aligned camera positions are therefore given by
\begin{equation}
\label{eq:scaled_camera_positions}
   t'_{0,j} = sRt_{0,j} + t.
\end{equation}


\subsection{Triangulation} 
\label{section:sign_positioning}
Finally, 
we compute the 3D traffic sign position through triangulation.
For each sign observed in a track of frames, the initial estimate of position is computed through the mid-point algorithm~\cite{szeliski2010computer}. In this approach, the coordinates ($c_{i,j}$) of sign $i$ in frame $j$ are transformed to directional vectors using the rectified camera intrinsics. Then, using linear least squares, the initial sign position is computed to minimize the distance to all directional vectors. 
Next, applying non-linear Bundle Adjustment (BA), the initial sign position estimate is refined by minimizing the reprojection error. Therefore, the absolute sign position
\begin{equation}
\label{eq:ba}
    p_i^{abs} = \underset{p_i}{\text {arg min}} \left(\sum_j\lVert K (R_{j,0}p_i + t'_{j,0}) - c_{i,j}\rVert ^2\right).
\end{equation}

We can either use the complete trajectory for triangulation of the sign positions, or use only those sub-sequences where the sign was observed. We compare the impact of using the full and short sequences on the accuracy of sign triangulation in section \ref{3d_triangulation}.

Thereafter, given the absolute sign positions, the relative positions $p_{i,j}^{rel}$ for each frame $j$ in which the sign was observed can be calculated. 
\begin{equation}
\label{eq:rel_traingulation_positioning}
    p_{i,j}^{rel} = R_{j,0}p_i^{abs} + t'_{j,0}.
\end{equation}
If the relative depth for any sign is calculated to be negative, we consider it to be a failed triangulation and discard it. Thereafter, the Mercator projection assumption is used to convert the estimated absolute traffic sign positions to the corresponding latitudes and longitudes.

%% file: experiments.tex
\section{Experiments}
In this section, we demonstrate the necessity of good self-calibration and the quantitative validity of our approach to single journey 3D traffic sign positioning without prior knowledge of camera parameters including distortion.

\subsection{Dataset}
Previous works on traffic sign positioning measured their accuracy using closed source datasets~\cite{welzel2014accurate, dabeer2017end}. Instead, we measure the 3D positioning accuracy of our approach against ground truth (GT) traffic sign positions\footnote{\url{https://github.com/hemangchawla/3d-groundtruth-traffic-sign-positions.git}\label{dataset}} in KITTI that we make publicly available to facilitate further research. This dataset was created using the matched images and LiDAR scans for sequences (Seq) 00 to 10 in the KITTI raw dataset~\cite{geiger2012we} (except Seq 03 which is missing from the raw dataset).
Using the low-resolution distorted images and corresponding GPS positions from the 10 sequences, we apply the proposed approach (see Fig.~\ref{fig:SingleJourney}) to triangulate the relative and absolute positions of detected traffic signs.

\begin{figure}[!htbp]
\begin{center}
    \includegraphics[width=0.75\linewidth]{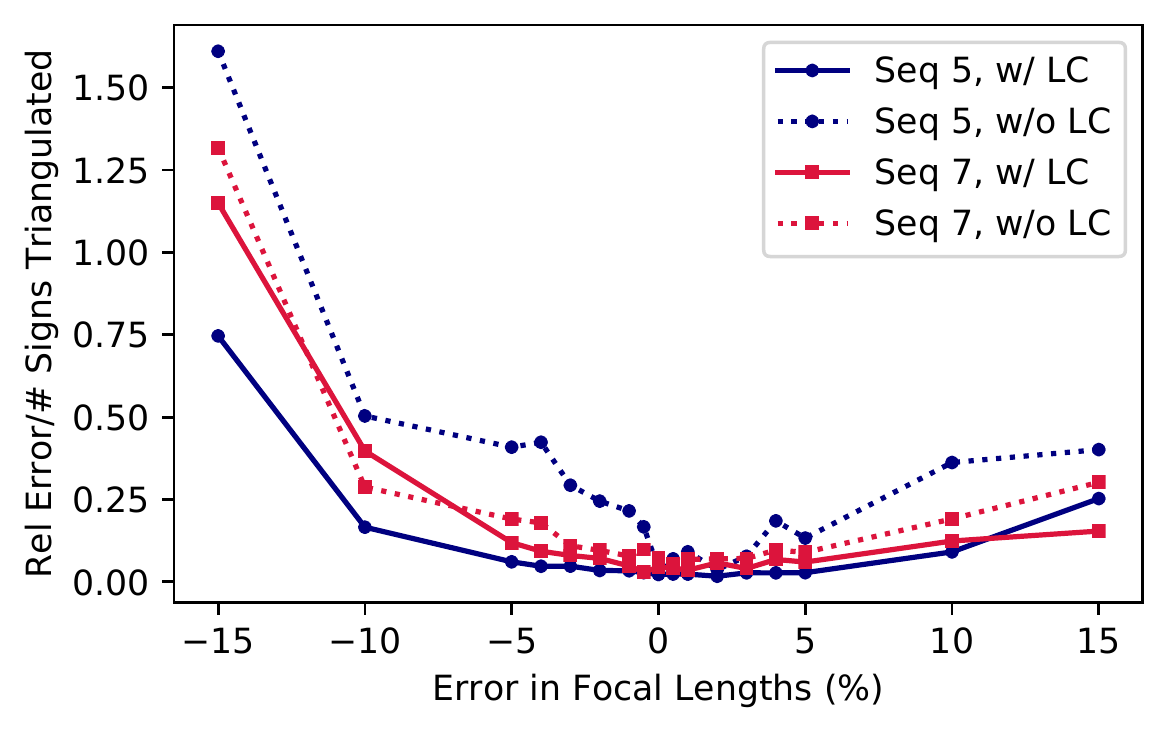}\\
    \includegraphics[width=0.75\linewidth]{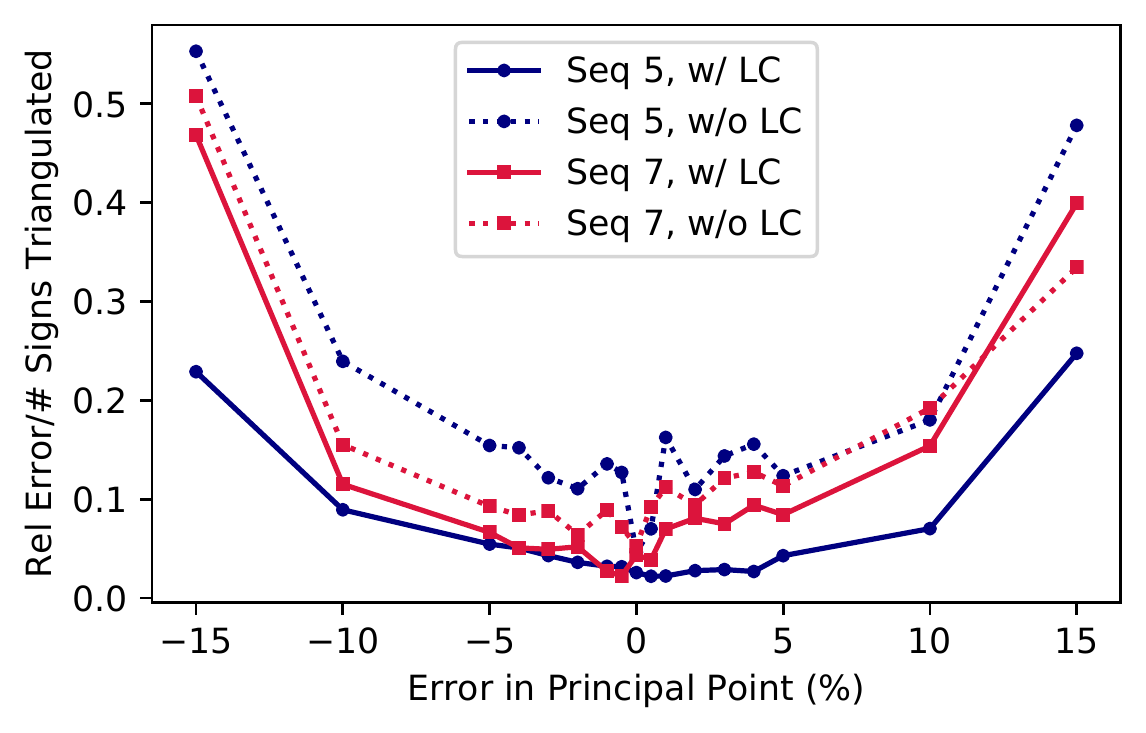}\\
    \includegraphics[width=0.75\linewidth]{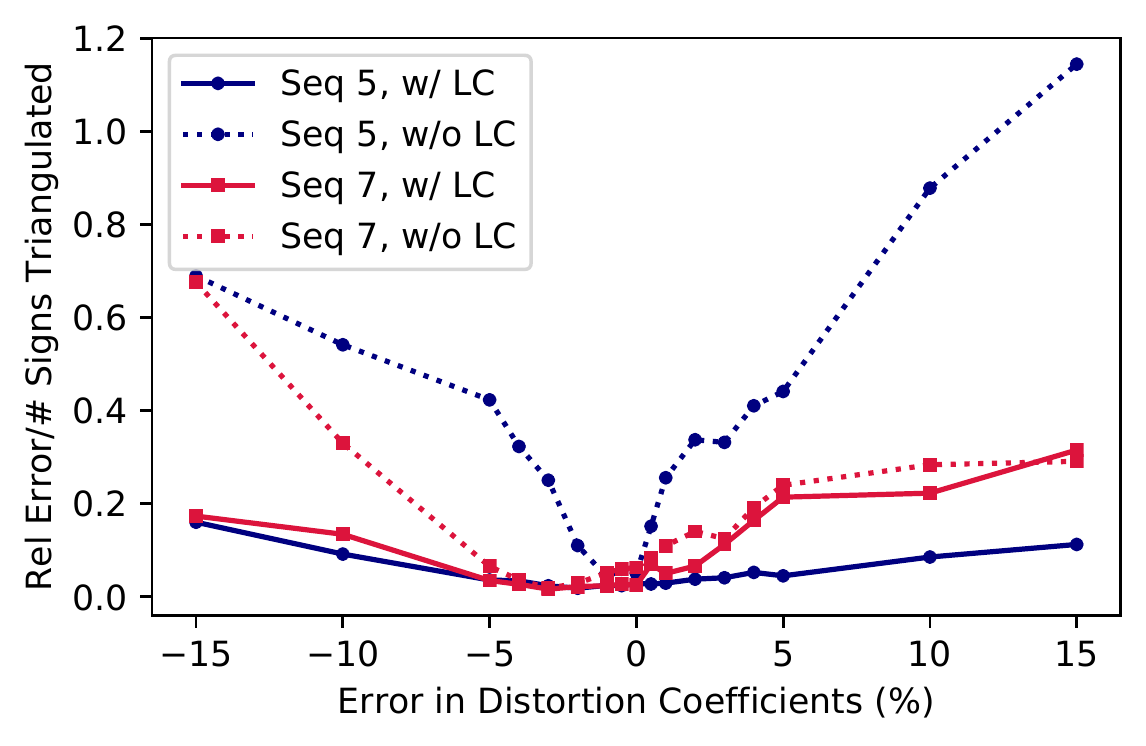}
\end{center}
  \caption{Individual sensitivity analysis. Top: performance for the error in focal lengths. Middle: performance for error in principal point. Bottom: performance for the error in distortion parameters.}
  \label{fig:OAT_sensitivity}
\end{figure}

\subsection{Sensitivity Analysis}
The existing approaches to 3D sign position triangulation assume known camera parameters. This requirement may not be met, or the parameters may change over time. In this section, we evaluate the impact of using incorrect camera focal lengths, principal point, and distortion parameters on positioning accuracy of traffic signs. In order to analyze the  sensitivity, we introduce error in GT focal length, principal point, and distortion coefficients. Using these incorrect parameters, we undistort the images and compute the rectified camera matrix. Using the rectified camera matrix, the ego-motion of the camera is computed through ORB-SLAM with (w/) and without (w/o) loop closure (LC). Thereafter, the traffic signs' positions are triangulated using the computed full camera trajectory. The performance with a chosen set of camera parameters is quantified as the average relative positioning error normalized by the number of successfully triangulated signs. For every set of parameters, we repeat the above experiment 10 times and report the corresponding minimum value.

\paragraph{Individual Sensitivity}
To evaluate the individual effects of using incorrect focal lengths, principal point, or distortion parameters, we perform the one-at-a-time sensitivity analysis. We measure the effect of introducing -15\% to +15\% error in one type of parameter while the others are set at their GT values. Fig. \ref{fig:OAT_sensitivity} shows the sensitivity of 3D positioning performance to the three types of camera parameters for KITTI Seq 05 (with multiple loops) and Seq 07 (with a single loop).
Note that for both sequences, when varying either the focal length, principal point, or the distortion coefficients, the performance is better w/ LC than w/o LC. Furthermore, with a higher number of loops (Seq 05) the difference of performance between the use of ORB-SLAM w/ and w/o LC is much higher. Specifically, observe that the performance is more sensitive to underestimating the focal length than overestimating it. However, the performance is equally sensitive to underestimating or overestimating the principal point. Also, the performance gap between sequence with multiple LC and the sequence with a single LC is higher when overestimating the distortion parameters.

\paragraph{Two-at-a-time}

\begin{figure}[tb!]
\begin{center}
    \includegraphics[width=\linewidth]{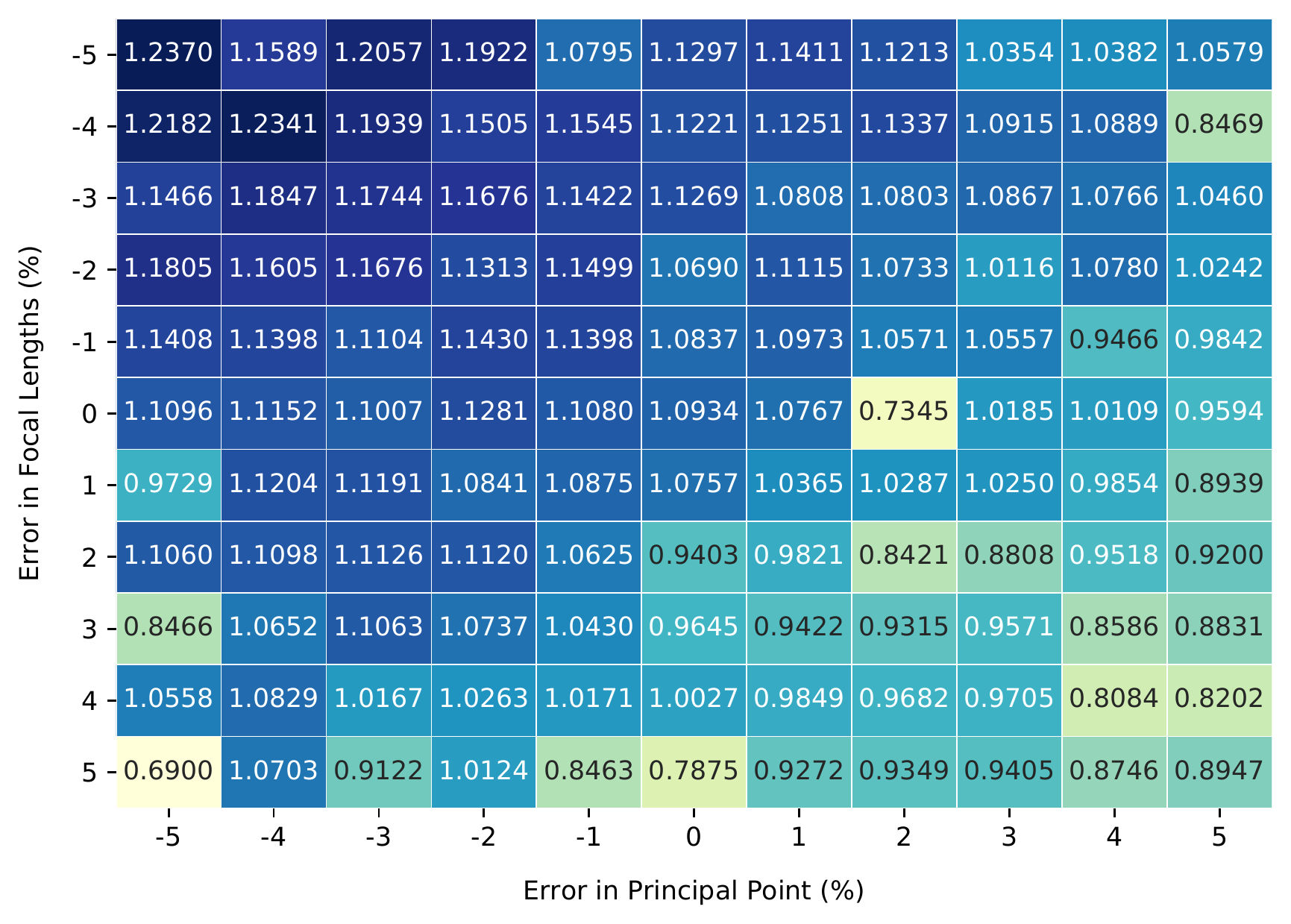}\\
    \includegraphics[width=\linewidth]{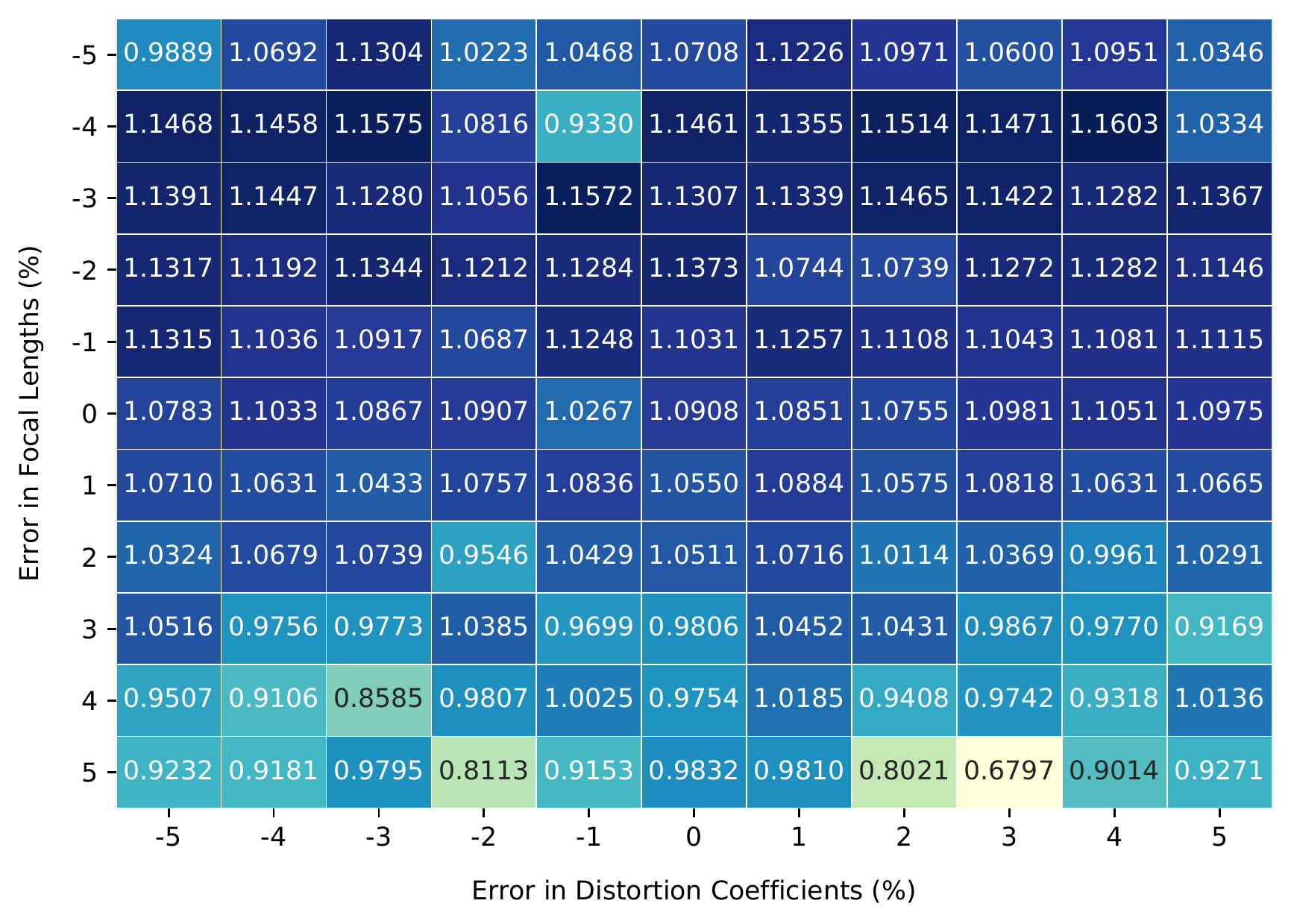}\\
    \includegraphics[width=\linewidth]{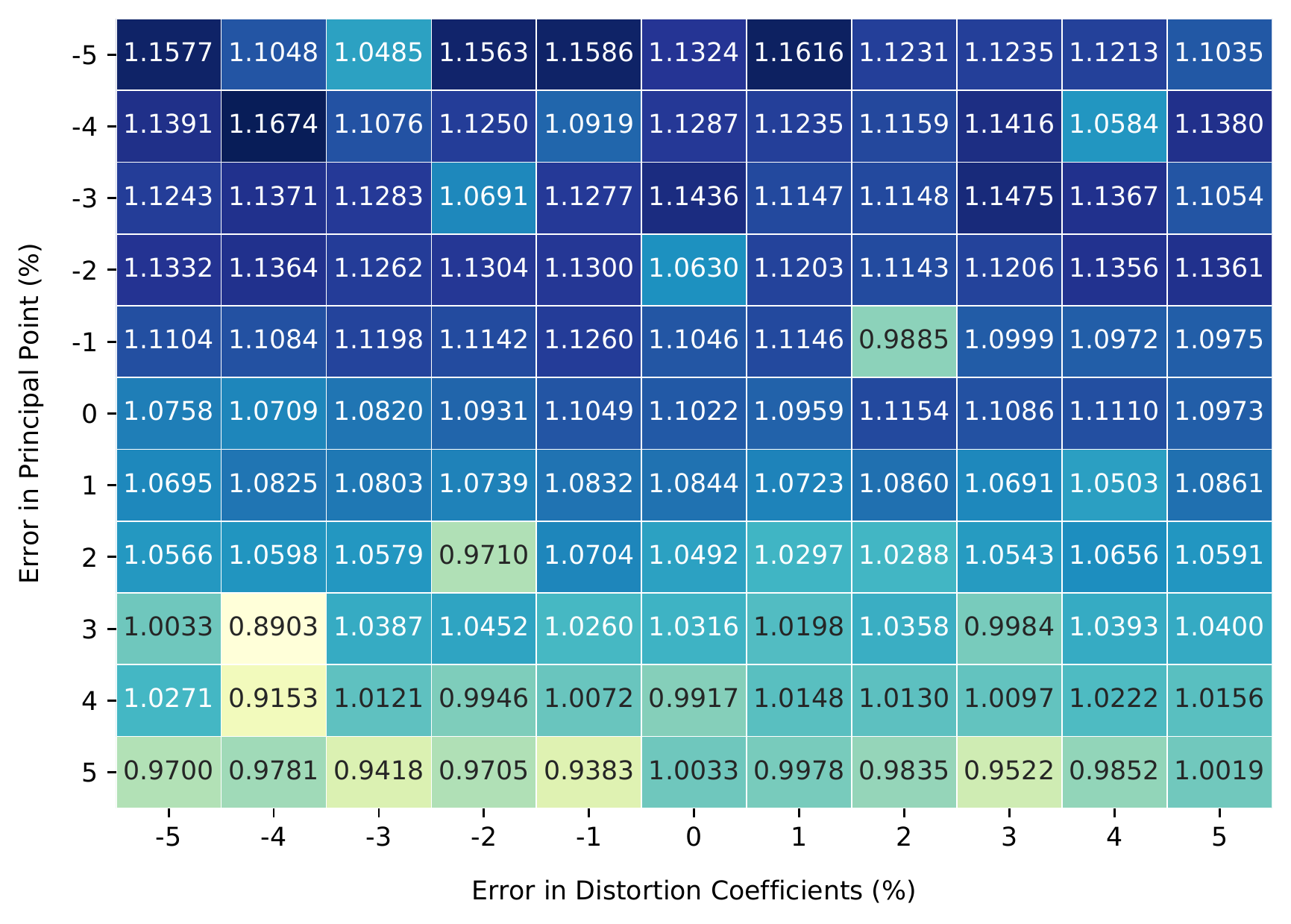}
\end{center}
  \caption{Two-at-a-time sensitivity analysis for Seq 05 (w/ LC). Top: Performance when varying focal lengths and principal point. Middle: Performance when varying focal length and distortion coefficients simultaneously. Bottom: Performance when varying principal point and distortion coefficients simultaneously.}
  \label{fig:int_sensitivity_5}
\end{figure}

\begin{figure}[tb!]
\begin{center}
    \includegraphics[width=\linewidth]{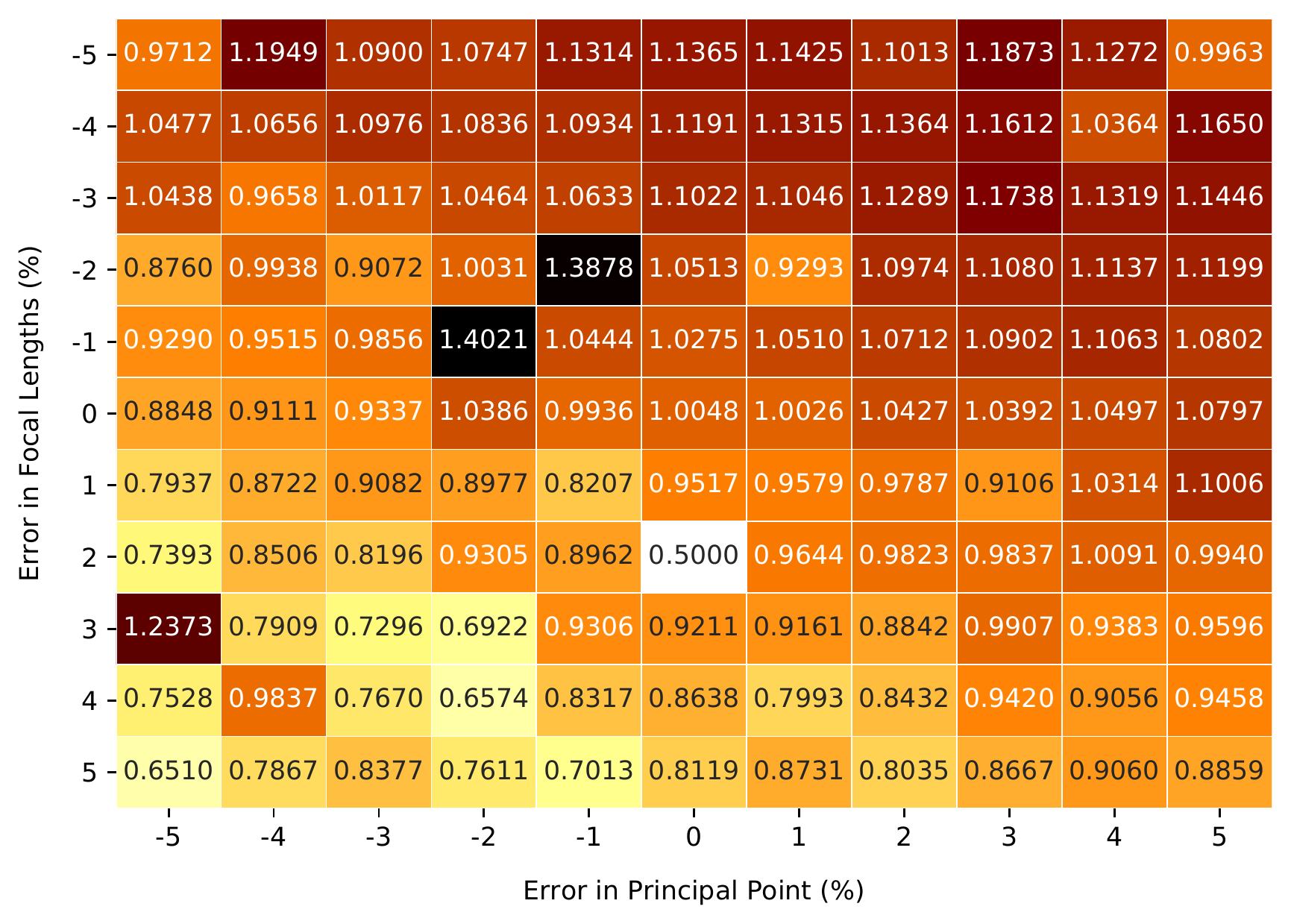}\\
    \includegraphics[width=\linewidth]{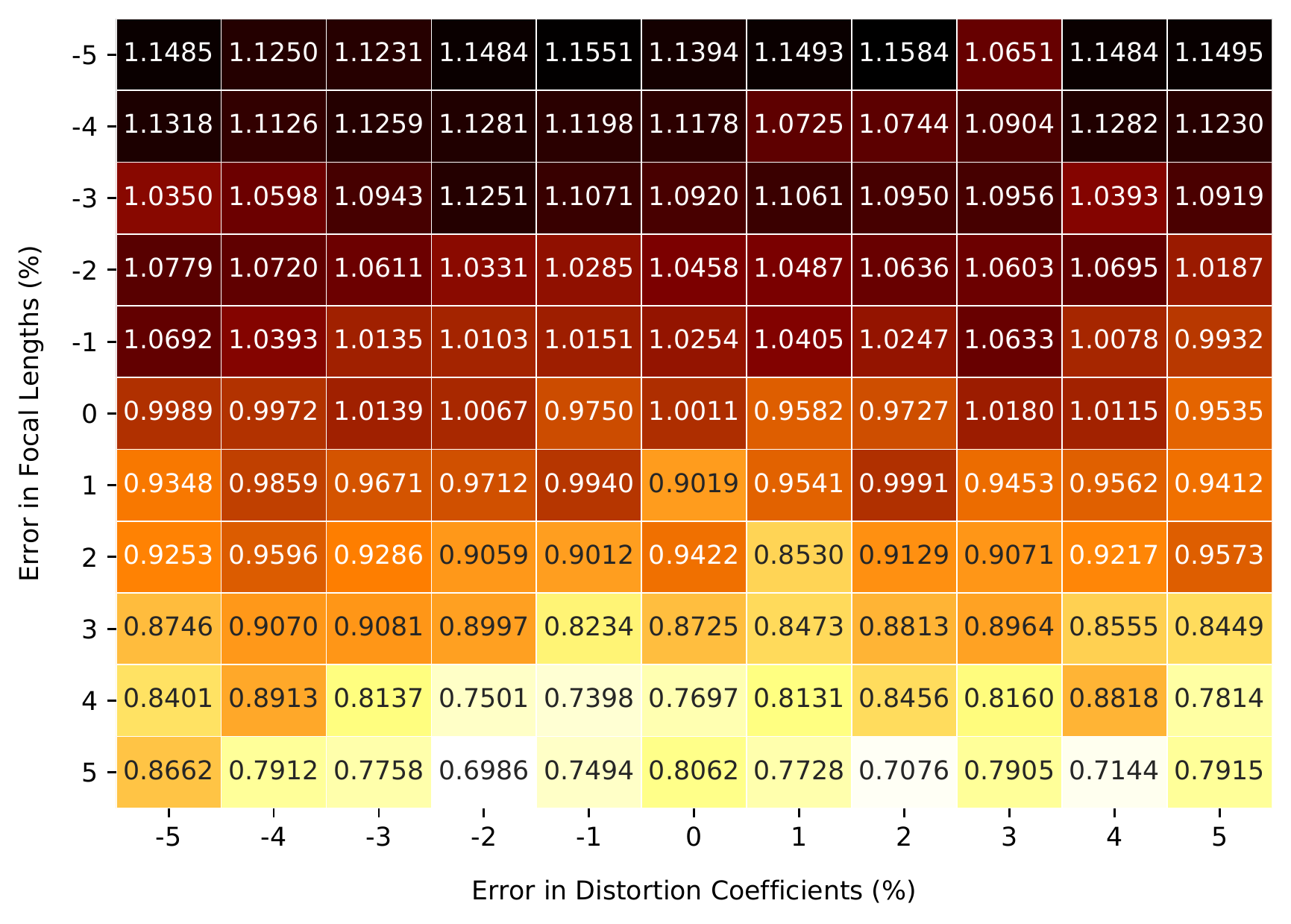}\\
    \includegraphics[width=\linewidth]{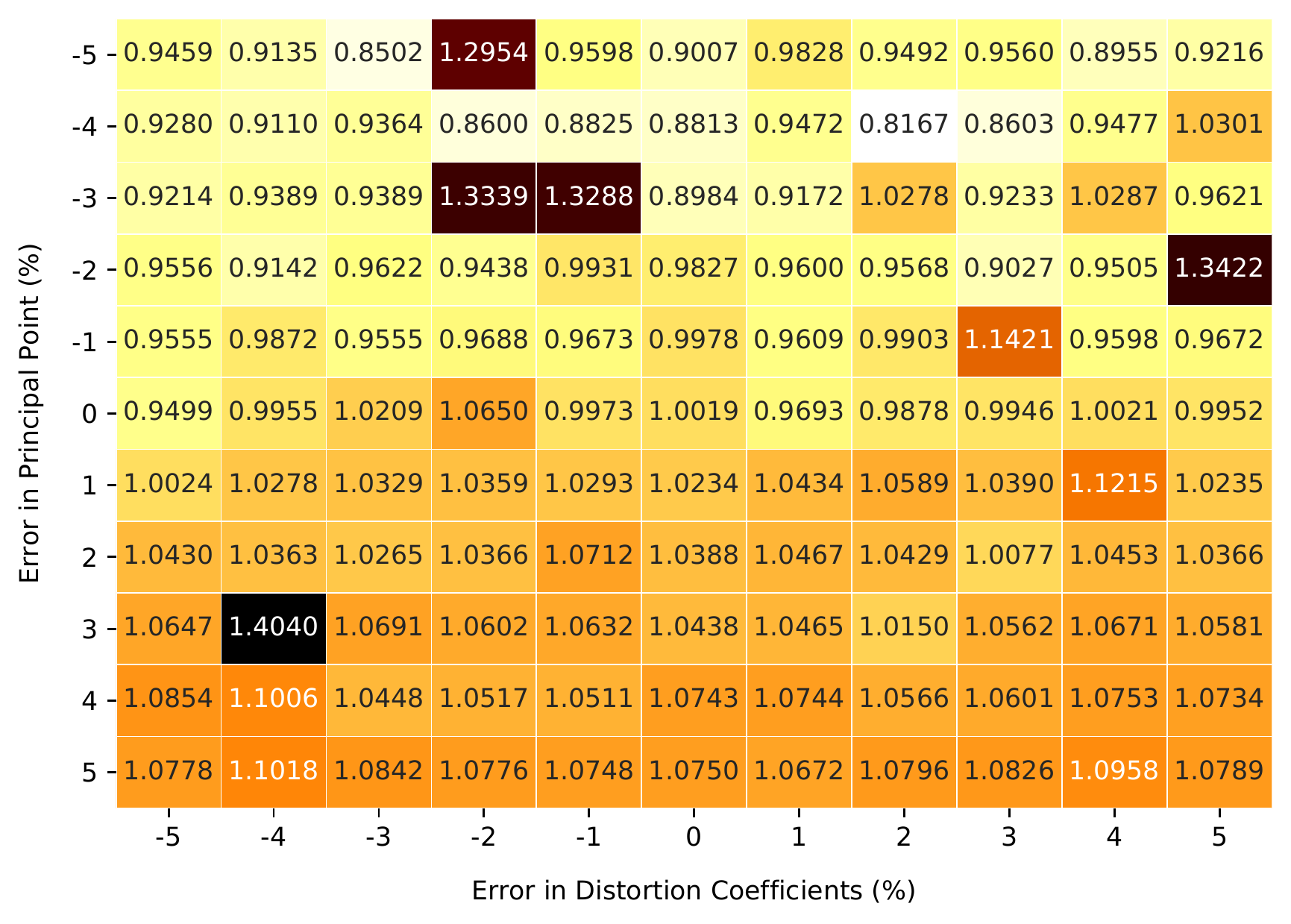}
\end{center}
  \caption{Two-at-a-time sensitivity analysis for Seq 07 (w/ LC). Top: Performance when varying focal lengths and principal point. Middle: Performance when varying focal length and distortion coefficients simultaneously. Bottom: Performance when varying principal point and distortion coefficients simultaneously.}
  \label{fig:int_sensitivity_7}
\end{figure}

Observe that overestimating the focal length is better than underestimating it, even when there are errors in the principal point or the distortion coefficients (cf. Figs. \ref{fig:int_sensitivity_5} and \ref{fig:int_sensitivity_7}). However, error in the principal point compensates for this and improves the performance. Specifically, overestimating and underestimating the principal point for Seq 05 and Seq 07 respectively improves the performance.
Similarly, when the focal length is incorrectly estimated, error in distortion coefficients compensate for it and improve the performance. This effect can be seen for both Seq 05 and 07. Moreover, the performance is more sensitive to the errors in the principal point than the distortion coefficients.

\begin{figure}[tb!]
\begin{center}
    \includegraphics[width=\linewidth]{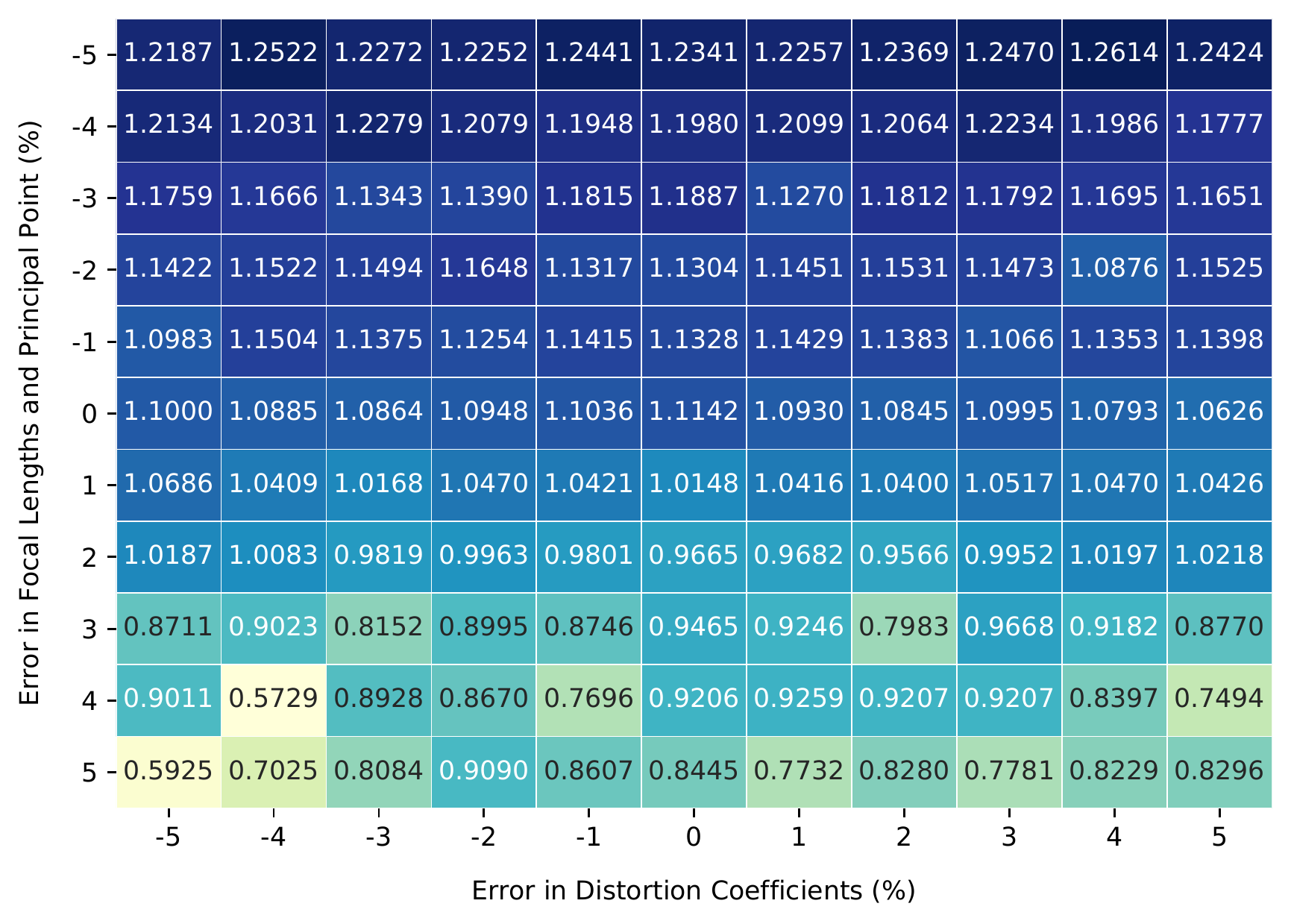}
\end{center}
  \caption{Interaction sensitivity analysis for Seq 05 (w/ LC). Performance when varying focal lengths and principal point simultaneously and independently from distortion coefficients.}
  \label{fig:int_sensitivity_all}
\end{figure}

Considering Figs. \ref{fig:int_sensitivity_5} and \ref{fig:int_sensitivity_7}, it can be seen that the performance is more sensitive to errors in the focal lengths and the principal point. This effect can also be seen when simultaneously varying the focal lengths and principal point against independently varying distortion coefficients for Seq 05 w/ LC, as shown in Fig. \ref{fig:int_sensitivity_all}. 

\subsection{Self-Calibration} 
We previously established that accurate self-calibration is important for good 3D traffic sign positioning. In this section, we quantify the accuracy of camera self-calibration with Colmap as part of the framework shown in Fig. \ref{fig:SingleJourney}. The GT camera parameters for Seq 00 to 02 are \{$f_x = 960.115, f_y = 954.891, c_x = 694.792, c_y = 240.355, \lambda_1 = -0.363, \lambda_2 = 0.151$\}, and the GT camera parameters for Seq 04 to 10 are \{$f_x = 959.198, f_y = 952.932, c_x = 694.438, c_y = 241.679, \lambda_1 = -0.369, \lambda_2 = 0.158$\}.

\begin{table}[tb!]
\caption{Self-Calibration Percentage Errors.}
\label{tab:self_calb}
\centering
\begin{tabular}{|c|cccccc|}
\hline
\textbf{Seq} & \boldsymbol{$\delta f_x$} & \boldsymbol{$\delta f_y$} & \boldsymbol{$\delta c_x$} & \boldsymbol{$\delta c_y$} & \boldsymbol{$\delta \lambda_1$} & \boldsymbol{$\delta \lambda_2$} \\ \hline
\textbf{00}   & 1.08 & -0.59 &  1.02 & 0.56  & -9.23  & -26.56 \\
\textbf{01}   & 1.31 & -5.77 & 0.22  & 4.07  & -8.71  & -19.81 \\
\textbf{02}   & 1.72 & 1.04  & 0.54  & 0.06  & -9.39  & -27.71 \\
\textbf{04}   & X    & X     & X     & X     & X      & X \\
\textbf{05}   & 0.95 & 0.72  & 0.35  & 1.07  & -10.21 & -28.57 \\
\textbf{06}   & 2.34 & 0.78  & 1.70  & -1.08 & -9.11  & -26.59 \\
\textbf{07}   & 2.19 & 1.28  & 0.51  & 1.50  & -9.89  & -28.16 \\
\textbf{08}   & 1.56 & 0.02  & 0.59  & 0.89  & -9.85  & -27.67 \\
\textbf{09}   & 2.14 & 4.45  & 0.80  & 0.61  & -5.46  & -21.01 \\
\textbf{10}   & 1.12 & 0.76  & 0.88  & 0.77  & -10.70 & -29.97 \\ \hline
\textbf{Avg}  & \bf 1.60 & \bf 0.30  & \bf 0.74  & \bf 0.94  & \bf -9.17  & \bf -26.23 \\ \hline
\end{tabular}
\end{table}

As elaborated in Table \ref{tab:self_calb}, the focal lengths are on average overestimated. While $f_x$ is overestimated when using any of the sequences, $f_y$ is overestimated for all except when using Seq 00 and 01, for self-calibration. Similarly, $c_x$ and $c_y$ are also overestimated on average. However, when using Seq 06 for self-calibration, $c_y$ is underestimated. While the percentage errors in the focal length and principal point are around 1\%, the percentage errors in estimating the distortion parameters are much higher. The signs for both the distortion coefficients are estimated correctly, negative for $\lambda_1$, and positive for $\lambda_2$. Furthermore, the percentage error in estimating $\lambda_2$, is higher than that for estimating $\lambda_1$. Note that Colmap fails to self-calibrate for Seq 04 because of the lack of any turns in that sequence.

\subsection{Ego-motion Estimation}
We also evaluate the absolute trajectory error (ATE) in meters for full~\cite{horn1987closed} and short 5-frame sequences~\cite{zhou2017unsupervised} using ORB-SLAM with GT calibration and Colmap to measure the effect of self-calibration on ego-motion estimation. Table \ref{table:ego_motion} shows the ego-motion performance for the 10 sequences considered from the KITTI dataset.

\begin{table}[!htbp]
\caption{Absolute Trajectory Error (ATE) for Ego-Motion Estimation with ORB-SLAM w/ and w/o Loop Closure. For each sequence top row uses GT calibration while the bottom row uses Colmap calibration.}
\label{table:ego_motion}
\centering
\begin{tabular}{|c|cc|cc|cc|}
\hline
\textbf{}   & \multicolumn{2}{c|}{\textbf{ATE full (\si\m)}} & \multicolumn{2}{c|}{\textbf{ATE-5 mean (\si\m)}} & \multicolumn{2}{c|}{\textbf{ATE-5 std (\si\m)}} \\ \hline
\textbf{Seq} & \textbf{w/ LC}    & \textbf{w/o LC}    & \textbf{w/ LC}     & \textbf{w/o LC}     & \textbf{w/ LC}     & \textbf{w/o LC}    \\ \hline
\multirow{2}{*}{\textbf{00}}   & 16.331  & 45.897  & 0.031 & 0.022 & 0.048 & 0.031 \\
                               & 12.320  & 144.749 & 0.023 & 0.009 & 0.050 & 0.013 \\ \hline
\multirow{2}{*}{\textbf{01}}   & X       & X       & X     & X     & X     & X \\
                               & X       & X       & X     & X     & X     & X \\ \hline
\multirow{2}{*}{\textbf{02}}   & 13.518  & 97.086  & 0.020 & 0.024 & 0.015 & 0.043 \\
                               & 32.020  & 159.555 & 0.010 & 0.009 & 0.011 & 0.008 \\ \hline
\multirow{2}{*}{\textbf{04}}   & 1.375   & 1.025   & 0.013 & 0.018 & 0.008 & 0.016 \\
                               & X       & X       & X     & X     & X     & X \\ \hline
\multirow{2}{*}{\textbf{05}}   & 4.876   & 29.093  & 0.012 & 0.009 & 0.019 & 0.007 \\
                               & 3.697   & 75.927  & 0.008 & 0.006 & 0.008 & 0.004 \\ \hline
\multirow{2}{*}{\textbf{06}}   & 14.112  & 50.904  & 0.012 & 0.011 & 0.008 & 0.010 \\
                               & 6.024   & 41.185  & 0.019 & 0.007 & 0.119 & 0.004 \\ \hline
\multirow{2}{*}{\textbf{07}}   & 3.194   & 16.272  & 0.013 & 0.009 & 0.018 & 0.006 \\
                               & 6.552   & 21.171  & 0.012 & 0.006 & 0.036 & 0.005 \\ \hline
\multirow{2}{*}{\textbf{08}}   & 45.575  & 40.787  & 0.011 & 0.011 & 0.011 & 0.010 \\
                               & 169.864 & 155.162 & 0.008 & 0.009 & 0.009 & 0.008 \\\hline
\multirow{2}{*}{\textbf{09}}   & 48.471  & 50.389  & 0.012 & 0.011 & 0.017 & 0.009 \\
                               & 30.330  & 37.019  & 0.008 & 0.008 & 0.006 & 0.006 \\\hline
\multirow{2}{*}{\textbf{10}}   & 5.856   & 7.230   & 0.008 & 0.008 & 0.006 & 0.006 \\
                               & 18.274  & 18.340  & 0.006 & 0.006 & 0.005 & 0.006 \\\hline\hline
\multirow{2}{*}{\textbf{Avg}}  & \bf 17.034  & \bf 37.631 & \bf 0.015 & \bf 0.014 & \bf 0.017 & \bf 0.015 \\
                               & \bf 35.135  & \bf 81.639 & \bf 0.012 & \bf 0.008 & \bf 0.031 & \bf 0.007 \\ \hline
\end{tabular}
\end{table}

Note that the ATE-full w/ LC is better than that w/o LC for both calibrations. However, ATE-5 mean and std are better w/o LC. While using Colmap for self-calibration slightly improves the ATE-5 mean implying better local agreement with the GT trajectory, the ATE-full is much worse implying poorer absolute localization. Since Colmap is unable to self-calibrate Seq 04 due to a lack of turns, its ego-motion estimation is not feasible. Also, Seq 01 suffers from tracking failure when using either of the calibrations.

\begin{table*}[!htbp]
\caption{Relative Errors in Traffic Sign Positioning using ORB-SLAM with full and short trajectories. Best results are highlighted in gray.}
\label{table:3d_comparison}
\centering
\begin{tabular}{|l|ccccc||ccccc|}
\hline
\textbf{}        & \multicolumn{5}{c||}{\textbf{ORB SLAM w/ LC}} & \multicolumn{5}{c|}{\textbf{ORB SLAM w/o LC}} \\\hline
\textbf{Seq}     & \textbf{$e_f$} & \textbf{$e_s$} & \textbf{$m$} & \textbf{$e_f/m$} & \textbf{$e_s/m$} & \textbf{$e_f$} & \textbf{$e_s$} & \textbf{$m$} & \textbf{$e_f/m$} & \textbf{$e_s/m$} \\ \hline
\textbf{00}      & 0.994 & 0.320 & 12& 0.083 & 0.027 & 6.099 & 0.276 & 12& 0.508 & 0.023 \\
\textbf{01}      & X     & X     & X & X     & X     & X     & X     & X & X     & X     \\
\textbf{02}      & 0.895 & 0.226 & 9 & 0.099 & 0.025 & 3.608 & 0.238 & 9 & 0.401 & 0.026 \\
\textbf{04}      & X     & X     & X & X     & X     & X     & X     & X & X     & X     \\
\textbf{05}      & 0.286 & 0.201 & 4 & 0.072 & 0.050 & 3.366 & 0.087 & 4 & 0.841 & 0.022 \\
\textbf{06}      & 0.311 & 0.235 & 2 & 0.156 & 0.118 & 3.551 & 0.336 & 2 & 1.776 & 0.168 \\
\textbf{07}      & 0.843 & 0.192 & 2 & 0.421 & 0.096 & 1.534 & 0.201 & 2 & 0.767 & 0.101 \\
\textbf{08}      & 3.547 & 0.330 & 5 & 0.709 & 0.066 & 2.711 & 0.332 & 5 & 0.542 & 0.066 \\
\textbf{09}      & 0.668 & 0.279 & 5 & 0.134 & 0.056 & 0.707 & 0.280 & 5 & 0.141 & 0.056 \\
\textbf{10}      & 0.692 & 0.146 & 3 & 0.231 & 0.049 & 0.683 & 0.099 & 3 & 0.228 & 0.033 \\ \hline
\textbf{Avg}     & \textbf{1.029} & \textbf{\colorbox{lightergray}{0.241}} & \textbf{5.250} & \textbf{0.238} & \textbf{\colorbox{lightergray}{0.061}} & \textbf{2.782} & \textbf{0.231} & \textbf{5.250} & \textbf{0.651} & \textbf{0.062} \\\hline
\end{tabular}
\end{table*}

\subsection{3D Traffic Sign Triangulation}
\label{3d_triangulation}
We evaluate the accuracy of crowdsourced 3D traffic sign positioning when triangulating using ORB-SLAM (w/ and w/o LC), and compare the effect of using short and full trajectories for triangulation (see Sec. \ref{section:sign_positioning}). In order to do so, we compute the mean relative error in sign positioning for all the sequences and normalize it by the number of signs successfully triangulated. 

Table \ref{table:3d_comparison} shows the mean relative errors using full ($e_f$) and short ($e_s$) trajectories. The number of signs triangulated is denoted by $m$. Note that Seq 01 is not triangulated due to the tracking failure of ORB-SLAM. Seq 04 is also not triangulated because of the failure to self-calibrate with Colmap. The best performance is given by the use of ORB-SLAM w/ LC through short trajectories. Better performance with short trajectories can be attributed to improved local scaling and alignment of estimated and GPS trajectories. Using loop closure also improves orientation due to global bundle adjustment, thereby resulting in more accurate triangulation. Therefore, it is preferable to triangulate the signs using only those sub-sequences where the sign was observed. A total of 42 traffic signs were successfully triangulated using the proposed method.

\begin{table*}[!htbp]
\caption{Mean Relative and Absolute 3D Traffic Sign Positioning Errors in meters on KITTI sequences.}
\label{table:3d_final}
\centering
\begin{tabular}{|l|cccccccccc|l|}
\hline
\textbf{Seq} & \textbf{00} & \textbf{01} & \textbf{02} & \textbf{04} & \textbf{05} & \textbf{06} & \textbf{07} & \textbf{08} & \textbf{09} & \textbf{10} & \textbf{Avg} \\\hline
\textbf{Rel} & 0.320 & X & 0.226 & X & 0.201 & 0.235 & 0.192 & 0.330 & 0.279 & 0.146 & 0.241 \\      
\textbf{Abs} & 1.246 & X & 1.178 & X & 0.309 & 0.536 & 1.134 & 4.022 & 0.983 & 0.949 & 1.295 \\ \hline
\end{tabular}
\end{table*}

The mean absolute triangulation errors are also computed for all the sequences, as shown in Table \ref{table:3d_final}. The average relative and absolute positioning error per sequence is \SI[mode=text]{0.241}{\m}, and \SI[mode=text]{1.295}{\m} respectively. The average absolute positioning error for all the signs is \SI[mode=text]{1.381}{\m}, while the relative positioning error over all frames is \SI[mode=text]{0.26}{\m}.

Our single journey \textit{relative} sign-positioning accuracy is comparable to that of \SI[mode=text]{0.46}{\m} accuracy achieved by Dabeer \etal ~\cite{dabeer2017end}. Unlike our work, it used known camera parameters as well as an IMU-GPS fusion for triangulation of 31 signs from the San Diego geological survey. 
Our single journey \textit{absolute} sign-positioning accuracy is comparable to that of \SI[mode=text]{1}{\m} accuracy achieved by Welzel \etal ~\cite{welzel2014accurate}. Unlike our work, it also relied upon prior knowledge of camera intrinsics and distortion coefficients, as well as  ground truth size and height of traffic signs for mapping 11 stop signs in Germany.

Note that we measure the positioning accuracy of traffic signs with different classes over multiple sequences of crowdsourced data. Hence, our results have a better representation of the various scenarios in which the traffic signs can be mapped using only a monocular camera and GPS, without prior knowledge of camera parameters.  

%% file: conclusion.tex
\section{Conclusion}

This work demonstrates monocular vision and GPS based crowdsourced mapping without knowing the camera focal lengths, principal point, and distortion coefficients a priori. 
Utilizing self-calibration, monocular ego-motion estimation, and triangulation in a single framework we accurately predict the 3D positions of traffic signs in a single journey. We also analyze the sensitivity of triangulation accuracy upon the accuracy of the camera parameters used. In the future, this accuracy may be improved by mapping through multiple journeys over the same path. We are also exploring deep learning based approaches for extending the map coverage to sequences in which multi-view-geometry based self-calibration and ego-motion estimation presently fail.